\def\year{2020}\relax
\documentclass[letterpaper]{article} % DO NOT CHANGE THIS
\usepackage{aaai20}  % DO NOT CHANGE THIS
\usepackage{times}  % DO NOT CHANGE THIS
\usepackage{helvet} % DO NOT CHANGE THIS
\usepackage{courier}  % DO NOT CHANGE THIS
\usepackage[hyphens]{url}  % DO NOT CHANGE THIS
\usepackage{graphicx} % DO NOT CHANGE THIS
\usepackage{graphicx}  % DO NOT CHANGE THIS
\usepackage{amsfonts}       % blackboard math symbols
\usepackage{amsmath}
\usepackage{mathabx}
\usepackage{subfig}
\usepackage{xcolor}
\usepackage{enumitem}
\usepackage{euscript}
\usepackage{soul}
\usepackage{tabulary}
\usepackage{multirow}
\usepackage{capt-of}
\usepackage{makecell}
\usepackage[ruled,linesnumbered]{algorithm2e}
\usepackage[capitalize]{cleveref}

\makeatletter 
\g@addto@macro{\@algocf@init}{\SetKwInOut{Parameter}{Parameters}} 
\makeatother

\newcommand{\alg}{\textsc{DIAL-GNN}\xspace}

\let\vec\mathbf

\newcommand{\citet}[1]{\citeauthor{#1} \shortcitep{#1}} \newcommand{\citep}{\cite}

\title{Deep Iterative and Adaptive Learning for Graph Neural Networks}
\author{Yu Chen\textsuperscript{\rm 1}, 
Lingfei Wu\textsuperscript{\rm 2}\thanks{Corresponding author.},
Mohammed J. Zaki\textsuperscript{\rm 1}\\\\
\textsuperscript{\rm 1}Rensselaer Polytechnic Institute\\ 
\textsuperscript{\rm 2}IBM Research\\ %If you have multiple authors and multiple affiliations
cheny39@rpi.edu, 
lwu@email.wm.edu, 
zaki@cs.rpi.edu
}

\begin{document}

\maketitle

\begin{abstract}

In this paper, we propose an end-to-end graph learning framework, namely \textbf{D}eep \textbf{I}terative and \textbf{A}daptive \textbf{L}earning for \textbf{G}raph \textbf{N}eural \textbf{N}etworks (\alg), for jointly learning the graph structure and graph embeddings simultaneously. We first cast the graph structure learning problem as a similarity metric learning problem and leverage an adapted graph regularization for controlling smoothness, connectivity and sparsity of the generated graph. We further propose a novel iterative method for searching for a hidden graph structure that augments the initial graph structure. Our iterative method dynamically stops when the learned graph structure approaches close enough to the optimal graph. Our extensive experiments demonstrate that the proposed \alg model can consistently outperform or match state-of-the-art baselines in terms of both downstream task performance and computational time. The proposed approach can cope with both transductive learning and inductive learning.

\end{abstract}

\section{Introduction}

% graph neural networks and applications
Recent years have seen a growing amount of interest in graph neural networks (GNNs)~\citep{kipf2016semi,li2016gated,GraphSage:hamilton2017inductive}, 
with successful applications in broad areas such as computer vision~\citep{norcliffe2018learning}, natural language processing~\citep{xu2018graph2seq,xu2018exploiting,xu2018sql} and healthcare informatics~\citep{gao2019dyngraph2seq}.  
Unfortunately, GNNs can only be used when graph-structured data is available. 
Many real-world applications naturally admit graph-structured data like social networks. However, it is questionable if these intrinsic graph-structures are optimal for the downstream tasks. 
More importantly, many applications such as those in natural language processing may only have non-graph structured data or even just the original feature matrix, requiring additional graph construction from the original data matrix to formulate graph data.

In the field of graph signal processing,
researchers have explored various ways of learning graphs from data, but without considering the downstream tasks~\citep{dong2016learning,kalofolias2016learn,kalofolias2017large,egilmez2017graph}. 
Independently, there has been an increasing amount of work studying the dynamic model of interacting systems utilizing implicit interaction models~\citep{sukhbaatar2016learning,hoshen2017vain,van2018relational,kipf2018neural}. 
However, these methods cannot be directly applicable to jointly learning the graph structure and graph representations when the graph is noisy or even not available. 
Recently, researchers have explored methods to automatically construct a graph~\citep{choi2019graph,li2018adaptive,liu2018contextualized,chen2019graphflow,chen2019reinforcement} 
when applying GNNs to non-graph structured data.
However, these methods merely optimize the graphs towards the  downstream tasks without utilizing the techniques which have proven to be useful in graph signal processing.

More recently, \citep{franceschi2019learning} presented a new approach for jointly learning the graph and the parameters of GNNs by approximately solving a bilevel program.  
%TODO: add back
% where they learnt a discrete probability distribution on the edges of the graph by approximately solving a bilevel program. 
% Their experimental results have shown promising performance in both cases where the input graph is either corrupted or not available. 
However, this approach has severe scalability issue since it needs to learn $N^2$ number of (Bernoulli) random variables to model joint probability distribution on the edges of the graph consisting of $N$ number of vertices. 
More importantly, it can only be used for transductive setting, which means this method cannot consider new nodes during the testing.

% Our proposed model DIAL-GNN 
To address these limitations, in this paper, we propose a \textbf{D}eep \textbf{I}terative and \textbf{A}daptive \textbf{L}earning for \textbf{G}raph \textbf{N}eural \textbf{N}etworks (\alg) framework for jointly learning the graph structure and the GNN parameters that are optimized towards some prediction task. 
% Our DIAL-GNN framework consisting of five parts: 1) a graph learning neural network to generate a graph topology; 2) a graph regularization neural network for controlling the smoothness, connectivity and sparsity of the generated graph; 3) a graph embedding neural network for generating node embeddings; 4) an iterative method to dynamically stop learning when the optimal graph is found; and 5) a prediction neural network for performing a downstream prediction task. 
In particular, we present a graph learning neural network that casts a graph learning problem as a data-driven similarity metric learning task for constructing a graph. 
We then adapt techniques for learning graphs from smooth signals \citep{kalofolias2016learn} to serve as graph regularization. 
More importantly, we propose a novel iterative method 
% to refine the graph structure with updated graph embeddings, and in the meanwhile, refine the graph embeddings with the updated graph structure.
to search for a hidden graph structure that augments the initial graph structure towards an optimal graph for the (semi-)supervised prediction tasks. 
The proposed approach can cope with both transductive learning and inductive learning.  
Our extensive experiments demonstrate that our model can consistently outperform or match state-of-the-art baselines on various datasets.
% in terms of both downstream task performance and computational time.

% We highlight three contributions of our approach as follows:
% \begin{itemize}
%     \item We propose an end-to-end graph learning framework for jointly learning graph structure and graph embedding simultaneously. The proposed approach can cope with both transductive training and inductive training.  
    
%     \item We propose to iteratively learn a better graph structure with updated node embeddings, and in the meanwhile, learn better node embeddings with the updated graph structure.
%     In addition, the iterative method dynamically stops when the learned graph structure approaches close enough to the optimal graph based on our proposed stopping criterion. 

%     \item We cast the graph structure learning problem as a similarity metric learning problem and leverage an adapted graph regularization for controlling smoothness, connectivity and sparsity of the generated graph.
    
%     \item Our extensive experiments demonstrate that our model can consistently outperform or match state-of-the-art baselines.
%     % TODO: add back
%     % in terms of both prediction performance on downstream tasks and computational time. 
% \end{itemize}

\section{Approach}

\begin{figure*}[!htb]
  \centering
    \includegraphics[keepaspectratio=true,scale=0.22]{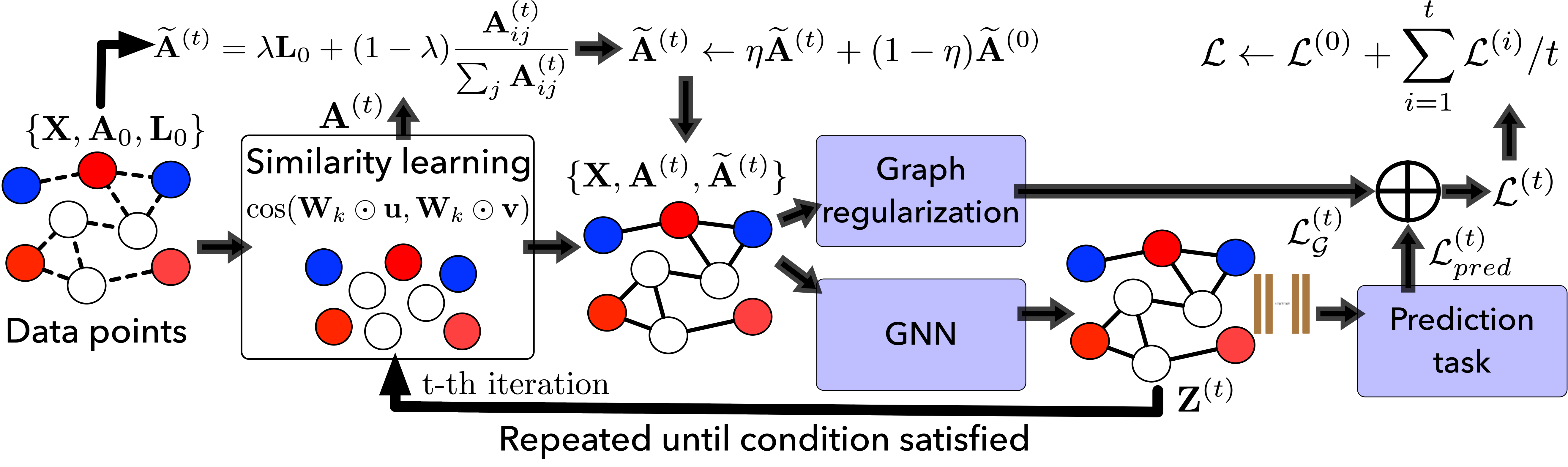}
  \caption{Overview of the proposed model. Dashed lines in the leftmost data points indicate the initial graph topology $\vec{A}_0$ either from the ground-truth graph if it exists or otherwise from the graph constructed using the kNN strategy. Best viewed in color. }
  \label{fig:overall_arch}
\end{figure*}

With this paper we address the challenging problem of automatic graph structure learning for GNNs.
We are given a set of $n$ objects $V$ associated with a feature matrix $\vec{X} \in \mathbb{R}^{d\times n}$ encoding the feature descriptions of the objects.
The goal is to automatically learn the graph structure $\mathcal{G}$, typically in the form of an adjacency matrix $\vec{A} \in \mathbb{R}^{n\times n}$, underlying the set of objects, which will be consumed by a GNN-based model for a downstream prediction task.
Unlike most existing methods that construct graphs based on hand-crafted rules or features during preprocessing,
our proposed \alg framework formulates the problem as an iterative learning problem that jointly learns the graph structure and the GNN parameters iteratively in an end-to-end manner. The overall model architecture is shown in Fig. \ref{fig:overall_arch}.

\subsection{Graph Learning as Similarity Metric learning}

A common strategy of graph construction is to first compute the similarity between pairs of nodes based on some metric, and then consume the constructed graph in a downstream task.
Unlike these methods, in this work, we design a learnable metric function for graph structure learning, which will be jointly trained with a task-dependent prediction model.

%TODO: add back
% In traditional graph theory, various methods have been explored to construct a graph from data points. These methods usually apply a metric function to compute the similarity between pairs of nodes during preprocessing, and then consume the constructed graph in a downstream task.
% Unlike these methods, in this work, we design a learnable metric function for graph structure learning, which will be jointly trained with a task-dependent prediction model.

\subsubsection{Similarity Metric Learning}

% Common options for metrics include cosine similarity, radial basis function (RBF) kernel and attention mechanisms.
% A good metric function is supposed to be learnable and expressively powerful.
After preliminary experiments, we design a multi-head weighted cosine similarity, %defined as,
\begin{equation}\label{eq:similarity_metric_learning}
% \vspace{-1mm}
\begin{aligned}
s_{ij}^k &= \text{cos}(\vec{w}_k \odot \vec{v}_i, \vec{w}_k \odot \vec{v}_j)\\
s_{ij} &= \frac{1}{m}\sum_{k=1}^{m}{s_{ij}^k}
\end{aligned}
\end{equation}
where $\odot$ denotes the Hadamard product.
Specifically, we use $m$ weight vectors (each has the same dimension as the input vectors and represents one perspective) to compute $m$ cosine similarity matrices independently and take their average as the final similarity $\vec{S}$.
Intuitively, $s_{ij}^k$ computes the cosine similarity between the two input vectors $\vec{v}_i$ and $\vec{v}_j$, for the $k$-th perspective where each perspective considers one part of the semantics captured in the vectors.
This idea of multi-head similarity is similar to those in multi-head attention~\citep{vaswani2017attention,velivckovic2017graph}.

\subsubsection{Graph Sparsification via $\varepsilon$-neighborhood}

An adjacency matrix (same for a metric) is supposed to be non-negative while $s_{ij}$ ranges between $[-1, 1]$.
In addition, many underlying graph structures are much more sparse than a fully connected graph, which is not only computationally expensive but also makes little sense for most applications.
We hence proceed to extract a symmetric sparse adjacency matrix $\vec{A}$ from $\vec{S}$ by considering only the $\varepsilon$-neighborhood for each node.
Specifically, we mask off those elements in $\vec{S}$ which are smaller than certain non-negative threshold $\varepsilon$.
\begin{equation}\label{eq:epsilon_neigh}
% \vspace{-1mm}
\begin{aligned}
\vec{A}_{ij} = 
\left\{
        \begin{array}{ll}
             s_{ij} & \quad  s_{ij} > \varepsilon  \\
              0 & \quad \text{otherwise}
        \end{array}
    \right.
\end{aligned}
\end{equation}

\subsection{Graph Regularization}

In graph signal processing~\citep{shuman2013emerging}, 
a widely adopted assumption for graph signals is that values change smoothly across adjacent nodes.
Given an undirected graph with symmetric weighted adjacency matrix $A$, the smoothness of a set of $n$ graph signals $\vec{x}_1, \dots, \vec{x}_n \in \mathbb{R}^{d}$ is usually measured by the Dirichlet energy~\citep{belkin2002laplacian}, 
\begin{equation}\label{eq:smoothness_loss}
% \vspace{-2mm}
\begin{aligned}
\Omega(\vec{A}, \vec{X})=\frac{1}{2n^2}\sum_{i,j}\vec{A}_{ij}||\vec{x}_i - \vec{x}_j||^2=\frac{1}{n^2}\text{tr}(\vec{X}^T \vec{L} \vec{X})
\end{aligned}
\end{equation}
where $\text{tr}(\cdot)$ denotes the trace of a matrix,
$\vec{L}=\vec{D}-\vec{A}$ is the graph Laplacian, and $\vec{D}=\sum_{j}\vec{A}_{ij}$ is the degree matrix. 
As can be seen, minimizing $\Omega(\vec{A}, \vec{X})$ forces adjacent nodes to have similar features, thus enforces
smoothness of the graph signals on the graph associated to $\vec{A}$.
However, solely minimizing the above smoothness loss will result in the trivial solution $\vec{A}=0$.
Also, it is desirable to have control of how sparse the resulting graph is.
Following~\citep{kalofolias2016learn},
we impose additional constraints to the learned graph,
\begin{equation}\label{eq:degree_sparsity_loss}
% \vspace{-2mm}
\begin{aligned}
f(\vec{A})= \frac{-\beta}{n} \vec{1}^T \text{log}(\vec{A}\vec{1}) +  \frac{\gamma}{n^2} ||\vec{A}||_F^2
\end{aligned}
\end{equation}
where $||\cdot||_F$ denotes the Frobenius norm.
As we can see, the first term penalizes the formation of disconnected graphs via the logarithmic barrier,
and the second term controls sparsity by penalizing large degrees due to the first term.

In this work, we borrow the above techniques, and apply them as regularization to the graph learned by~\cref{eq:similarity_metric_learning,eq:epsilon_neigh}.
The overall graph regularization loss is defined as the sum of the above losses, which is able to control the smoothness, connectivity and sparsity of the resulting graph where $\alpha$, $\beta$ and $\gamma$ are all non-negative hyperparameters.
\begin{equation}\label{eq:graph_loss}
\begin{aligned}
\mathcal{L}_{\mathcal{G}} = \alpha   \Omega(\vec{A}, \vec{X}) + f(\vec{A})
\end{aligned}
\end{equation}
% Note that $\alpha$, $\beta$ and $\gamma$ are all non-negative hyperparameters.

% \subsection{Iterative Method for Joint Graph Structure and Representation Learning}

\subsection{An Iterative Graph Learning Method}

\subsubsection{Joint Graph Structure and Representation Learning}

We expect the graph structure underlying a set of objects to serve two purposes:
i)
it should respect the semantic relations among the objects, which is enforced by the metric function (\cref{eq:similarity_metric_learning}) and the smoothness loss (\cref{eq:smoothness_loss});
ii),
it should suit the needs of the downstream prediction task.
Compared to previous works which directly optimize the adjacency matrix based on either some graph regularization loss~\citep{kalofolias2017large}, or some task-dependent prediction loss~\citep{franceschi2019learning},
% we propose to learn an optimal similarity metric function as well as the GNN parameters by  minimizing a joint loss function combining both the task prediction loss and the graph regularization loss, namely, $\mathcal{L} = \mathcal{L}_{\text{pred}} + \mathcal{L}_{\mathcal{G}}$.
we propose to learn by minimizing a joint loss function combining both the task prediction loss and the graph regularization loss, namely, $\mathcal{L} = \mathcal{L}_{\text{pred}} + \mathcal{L}_{\mathcal{G}}$.

Note that our graph learning framework is agnostic to various GNNs and prediction tasks.
In this paper,
we adopt a two-layered GCN~\citep{kipf2016semi} where the first layer maps the node features to the node embedding space (\cref{eq:update_node_vec}), and the second layer further maps the intermediate node embeddings to the output space (\cref{eq:prediction}). 
\begin{align}
% \vspace{-4mm}
  & \vec{Z} = \text{ReLU}(\widetilde{\vec{A}} \vec{X} \vec{W}_1) \label{eq:update_node_vec}\\
  & \widehat{\vec{y}} = \sigma(\widetilde{\vec{A}} \vec{Z} \vec{W}_2) \label{eq:prediction}\\
  & \mathcal{L}_{\text{pred}} = \ell(\widehat{\vec{y}}, \vec{y}) \label{eq:pred_loss}
\end{align}
where $\widetilde{\vec{A}}$ is the normalized adjacency matrix, $\sigma(\cdot)$ is a task-dependent output function, and $\ell(\cdot)$ is a task-dependent loss function.
For instance, for node classification problem,
$\sigma(\cdot)$ is a softmax function for predicting a probability distribution over a set of classes,
% and $\vec{W}_2 \in \mathbb{R}^{h \times |y|}$ where $|y|$ is the number of node labels.
and $\ell(\cdot)$ is a cross-entropy function for computing the prediction loss.

We now discuss how to obtain the normalized adjacency matrix $\widetilde{\vec{A}}$.
% For some problems when an initial graph is available, 
Our preliminary experiments showed that it is harmful to totally discard the initial graph structure when it is available.
Previous works~\citep{velivckovic2017graph,jiang2019semi} inject the initial graph structure into the graph learning mechanism by performing masked attention,
which might limits its graph learning ability.
This is because there is no way for their methods to learn weights for those
edges that do not exist in the initial graph, but carry useful topological information.
With the assumption that the optimal graph structure is potentially a small shift from the initial graph structure,
we combine the learned graph structure with the initial graph structure as follows, 
\begin{equation}\label{eq:adj_norm}
% \vspace{-2mm}
\begin{aligned}
\widetilde{\vec{A}} = \lambda \vec{L}_0 + (1 - \lambda) \frac{\vec{A}_{ij}}{\sum_j{\vec{A}_{ij}}}
\end{aligned}
\end{equation}
where $\vec{L}_0$ is the normalized adjacency matrix of the initial graph, defined as,
$\vec{L}_0 = \vec{D}_0^{-1/2}\vec{A}_0\vec{D}_0^{-1/2}$,
and $\vec{D}_0$ is its degree matrix.
The adjacency matrix learned by~\cref{eq:similarity_metric_learning,eq:epsilon_neigh} is row normalized such that each row sums to 1.
A hyperparameter $\lambda$ is used to balance the trade-off between the learned graph structure and the initial graph structure.
If such an initial graph structure is not available, we instead use a kNN graph constructed based on cosine similarity.

\begin{algorithm}[!htb]
\footnotesize
\caption{\alg}
\label{alg:DIAL-GNN}
\DontPrintSemicolon
\SetAlgoNoLine
 \KwIn{$\vec{X}$, $\vec{y}$$[, \vec{A}_0]$}
 \Parameter{$m$, $\varepsilon$, $\alpha$, $\beta$, $\gamma$, $\lambda$, $\delta$, $T$, $\eta$$[, k]$}
 \KwOut{$\Theta$, $\widetilde{\vec{A}}^{(t)}$, $\widehat{\vec{y}}$}
 $[\vec{A}_0 \leftarrow  \text{kNN}( \vec{X}, k)]$ 
 \tcp*{Init. $\vec{A}_0$ to kNN graph if $\vec{A}_0$ is unavailable}

    % Init.
     $\vec{A}^{(0)}, \widetilde{\vec{A}}^{(0)} \leftarrow \{\vec{X}, \vec{A}_0\}$ using ~\cref{eq:similarity_metric_learning,eq:epsilon_neigh,eq:adj_norm}
     \tcp*{Learn the adj. matrix}
     
    $\vec{Z}^{(0)} \leftarrow \{\widetilde{\vec{A}}^{(0)}, \vec{X}\}$ using~\cref{eq:update_node_vec}
    \tcp*{Compute node embeddings}

    % Compute loss-1
        $\mathcal{L}_{\text{pred}}^{(0)} \leftarrow \{\widetilde{\vec{A}}^{(0)}, \vec{Z}^{(0)}, \vec{y}\}$ using~\cref{eq:prediction,eq:pred_loss}
        \tcp*{Compute prediction loss}
      
        $\mathcal{L}_{\mathcal{G}}^{(0)} \leftarrow \{\vec{A}^{(0)}, \vec{X}\}$ using~\cref{eq:smoothness_loss,eq:degree_sparsity_loss,eq:graph_loss}
        \tcp*{Compute graph regularization loss}

        $\mathcal{L}^{(0)} \leftarrow \mathcal{L}_{\text{pred}}^{(0)} + \mathcal{L}_{\mathcal{G}}^{(0)}$
        \tcp*{Compute joint loss}

     $t \leftarrow 0$\;
    %  $\mathcal{L}^1 \leftarrow 0$\;
 \While{$(t == 0\quad \text{or} \quad  ||\vec{A}^{(t)} - \vec{A}^{(t-1)}||_F^2 > \delta ||\vec{A}^{(0)}||_F^2 ) \quad \text{and} \quad t < T$}{\label{alg_line:stopping_criterion}
      
    $t \leftarrow t + 1$\;
      
    $\vec{A}^{(t)}, \widetilde{\vec{A}}^{(t)} \leftarrow \{\vec{Z}^{(t-1)}, \vec{A}_0\}$ using~\cref{eq:similarity_metric_learning,eq:epsilon_neigh,eq:adj_norm}
    \label{alg_line:refine_adj}
    \tcp*{Refine the adj. matrix}

    % $\widetilde{\vec{A}}^{(t)} \leftarrow \eta \widetilde{\vec{A}}^{(t)} + (1 - \eta) \widetilde{\vec{A}}^{(0)}$\;
    % \label{alg_line:combine_adj_norm_t_0}

    $\widebar{\vec{A}}^{(t)} \leftarrow \{\widetilde{\vec{A}}^{(t)}, \widetilde{\vec{A}}^{(0)}\}$ using~\cref{eq:combine_adj_norm_t_0}
    \label{alg_line:combine_adj_norm_t_0}

    $\vec{Z}^{(t)} \leftarrow \{\widebar{\vec{A}}^{(t)}, \vec{X}\}$ using~\cref{eq:update_node_vec}
    \label{alg_line:refine_node_vec}
    \tcp*{Refine node embeddings}

    % Compute loss-2   
        $\widehat{\vec{y}} \leftarrow \{\widebar{\vec{A}}^{(t)}, \vec{Z}^{(t)}\}$ using~\cref{eq:prediction}
        \tcp*{Compute task output}

        $\mathcal{L}_{\text{pred}}^{(t)} \leftarrow \{\widehat{\vec{y}}, \vec{y}\}$ using~\cref{eq:pred_loss}\;

        $\mathcal{L}_{\mathcal{G}}^{(t)} \leftarrow \{\vec{A}^{(t)}, \vec{X}\}$ using~\cref{eq:smoothness_loss,eq:degree_sparsity_loss,eq:graph_loss}\;
        
        % $\mathcal{L}_{\Delta}^1 \leftarrow \{\vec{A}^{(t)}, \vec{A}^{(t-1)}\}$ using~\cref{eq:graph_delta}
        % \tcp*{Penalize aggressive graph update}

        $\mathcal{L}^{(t)} \leftarrow  \mathcal{L}_{\text{pred}}^{(t)} + \mathcal{L}_{\mathcal{G}}^{(t)}$\;\label{alg_line:joint_loss_t}
     }

    $\mathcal{L} \leftarrow \mathcal{L}^{(0)}  + \sum_{i=1}^t{\mathcal{L}^{(i)}} / t$\; 
    
    \If{Training}{\label{alg_line:BP_begin}
    
    Back-propagate $\mathcal{L}$ to update model weights $\Theta$\;\label{alg_line:BP}
    
    }\label{alg_line:BP_end}
    
    % $\widehat{\vec{y}} \leftarrow \{\widetilde{\vec{A}}^{(t)}, \vec{Z}^{(t)}\}$ using~\cref{eq:prediction}
    % \tcp*{Compute task output}
    %  \vspace{-1mm}
\end{algorithm}

\subsubsection{Iterative Method for Graph Learning}

Some previous works~\citep{velivckovic2017graph} rely solely on raw node features to learn the graph structure based on some attention mechanism, which we think have some limitations since raw node features might not contain enough information to learn good graph structures. 
% We observe that learning network structures entirely from node features can sometimes be challenging, especially for those datasets (e.g., Cora and Citeseer) where node features are sparse and not very distinguishable.
Our preliminary experiments showed that simply applying some attention function upon these raw node features does not help learn meaningful graphs (i.e., attention scores are kind of uniform).
Even though we train the model jointly using the task-dependent prediction loss, we are limited by the fact that the similarity metric is computed based on the potentially inadequate raw node features.
% To put it simply, if the raw node features are not very helpful,
% it is challenging to learn good graph structures from them.

To address the above limitation, 
we propose a Deep Iterative and Adaptive Learning framework for Graph Neural Networks (\alg).
A sketch of the \alg framework is presented in~\cref{alg:DIAL-GNN}. Inputs and operations in squared brackets are optional.
Specifically, besides computing the node similarity based on their raw features,
we further introduce another learnable similarity metric function (~\cref{eq:similarity_metric_learning}) that is rather computed based on the intermediate node embeddings, as demonstrated in~\cref{alg_line:refine_adj}.
% Compared to the raw node features, these intermediate node embeddings usually reside on a low-dimensional manifold of the raw node feature space, and are optimized towards the downstream prediction task.
The aim is that the metric function defined on this node embedding space is able to learn topological information supplementary to the one learned solely based on the raw node features.
In order to combine the advantages of both the raw node features and the node embeddings,
we make the final learned graph structure as a linear combination of them,
% , as shown in~\cref{alg_line:combine_adj_norm_t_0}.
\begin{equation}\label{eq:combine_adj_norm_t_0}
% \vspace{-2mm}
\begin{aligned}
\widebar{\vec{A}}^{(t)} = \eta \widetilde{\vec{A}}^{(t)} + (1 - \eta) \widetilde{\vec{A}}^{(0)}
\end{aligned}
\end{equation}
where $\widetilde{\vec{A}}^{(t)}$ and $\widetilde{\vec{A}}^{(0)}$ are the two normalized adjacency matrices learned by~\cref{eq:adj_norm} at the $t$-th iteration and the initialization step before the iterative loop, respectively.

Furthermore,
as we can see from~\cref{alg_line:refine_adj} to~\cref{alg_line:refine_node_vec}, the algorithm repeatedly refines the adjacency matrix $\widetilde{\vec{A}}^{(t)}$ with the updated node embeddings $\vec{Z}^{(t-1)}$, 
and in the meanwhile, refines the node embeddings $\vec{Z}^{(t)}$ with the updated adjacency matrix $\widetilde{\vec{A}}^{(t)}$.
The iterative procedure dynamically stops when the learned adjacency matrix converges (with certain threshold $\delta$) or the maximal number of iterations is reached (~\cref{alg_line:stopping_criterion}).
Compared to using a fixed number of iterations globally, the advantage of  applying this dynamical stopping strategy becomes more clear when we are doing mini-batch training since we can adjust when to stop dynamically for each example graph in the mini-batch. At each iteration, a joint loss combining both the task-dependent prediction loss and the graph regularization loss is computed (~\cref{alg_line:joint_loss_t}).
After all iterations,
the overall loss will be back-propagated through all previous iterations to update the model parameters (\cref{alg_line:BP}).
% Due to the space limit, we provide formal analysis on the convergence property and complexity of the model in
% ~\cref{sec:formal_analysis}.
% the appendix.

\section{Formal Analysis}\label{sec:formal_analysis}

\subsection{Convergence of the Iterative Learning Procedure}\label{sec:convergence_analysis}

\begin{figure}[!htb]
\vspace{-1mm}
  \centering
    \includegraphics[keepaspectratio=true,scale=0.18]{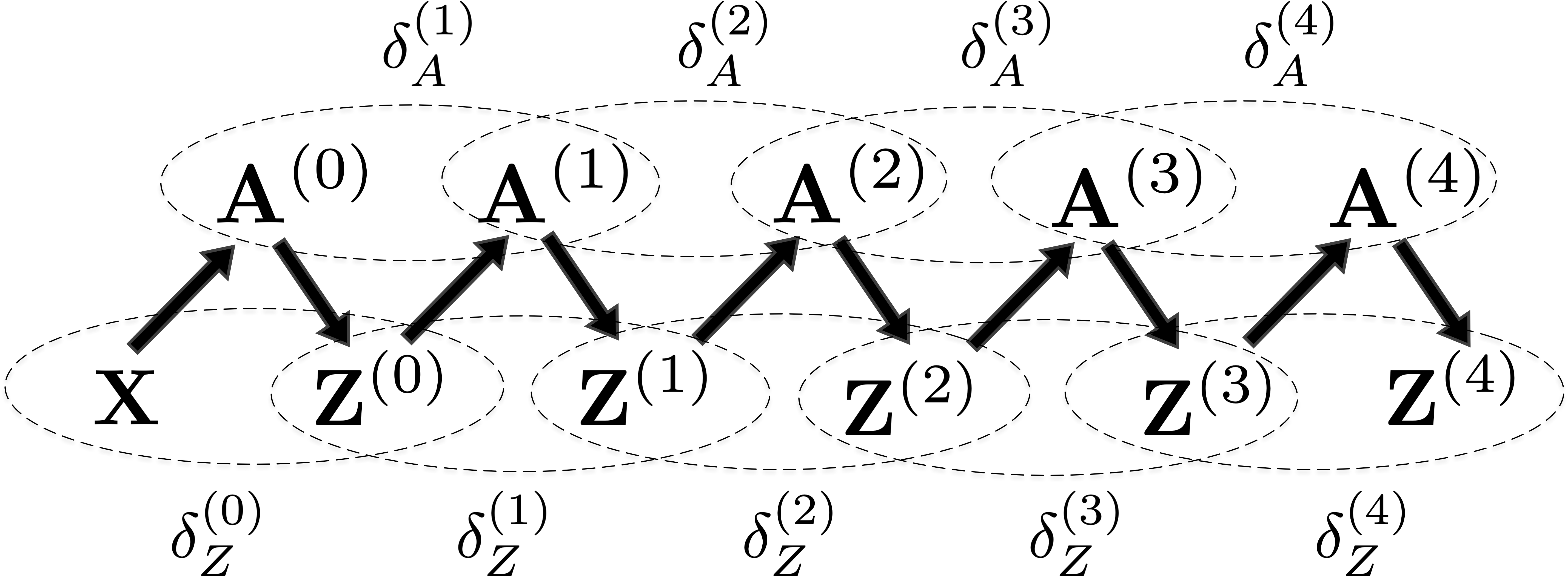}
  \caption{Information flow of the proposed iterative learning procedure.}
  \label{fig:info_flow}
\end{figure}

While it is challenging to theoretically prove the convergence of the proposed iterative learning procedure due to the arbitrary complexity of the involved learning model, here we want to conceptually understand why it works in practice.
\cref{fig:info_flow} shows the information flow of the learned adjacency matrix $\vec{A}$ and the intermediate node embedding matrix $\vec{Z}$ during the iterative procedure. 
For the sake of simplicity, we omit some other variables such as $\widetilde{\vec{A}}$.
As we can see, at $t$-th iteration, $\vec{A}^{(t)}$ is computed based on $\vec{Z}^{(t-1)}$ (\cref{alg_line:refine_adj}),
and $\vec{Z}^{(t)}$ is computed based on $\widetilde{\vec{A}}^{(t)}$ (\cref{alg_line:refine_node_vec}) which is computed based on $\vec{A}^{(t)}$ (\cref{eq:adj_norm}).
We further denote the difference between the adjacency matrices at the $t$-th iteration and the previous iteration by $\delta_A^{(t)}$.
Similarly, we denote the difference between the node embedding matrices at the $t$-th iteration and the previous iteration by $\delta_Z^{(t)}$.

If we assume that $\delta_Z^{(1)} < \delta_Z^{(0)}$, then we can expect that
$\delta_A^{(2)} < \delta_A^{(1)}$ because conceptually more similar node embedding matrix (i.e., smaller $\delta_Z$) is supposed to produce more similar adjacency matrix (i.e., smaller $\delta_A$) given the fact that model parameters keep the same through iterations.
Similarly, given that $\delta_A^{(2)} < \delta_A^{(1)}$, we can expect that  $\delta_Z^{(2)} < \delta_Z^{(1)}$.
Following this chain of reasoning, we can easily extend it to later iterations.
In order to see why the assumption $\delta_Z^{(1)} < \delta_Z^{(0)}$ makes sense in practice, we need to recall the fact that $\delta_Z^{(0)}$ measures the difference between $\vec{Z}^{(0)}$ and $\vec{X}$, which is usually larger than the difference between $\vec{Z}^{(1)}$ and $\vec{Z}^{(0)}$, namely $\delta_Z^{(1)}$.
We will empirically examine the convergence property of the iterative learning procedure in the experimental section.

\subsection{Model Complexity}

The cost of learning an adjacency matrix is $\mathcal{O}(n^2h)$ for $n$ nodes and data in $\mathbb{R}^h$, while computing node embeddings costs $\mathcal{O}(n^2d + ndh)$, 
computing task output costs $\mathcal{O}(n^2h)$, 
and computing the total loss costs $\mathcal{O}(n^2d)$.
We set the maximal number of iterations to $T$, hence
the overall complexity is $\mathcal{O}(Tn(nh+nd+hd))$.
If we assume that $d \approx h$ and $n \gg d$, the overall complexity is $\mathcal{O}(Tdn^2)$.

\section{Experiments}

\begin{table*}[!htb]
\vspace{-5mm}
\caption{Test accuracy ($\pm$ standard deviation) in percentage on various classification datasets in the transductive setting.}
\label{table:transductive_results}
\centering
\scalebox{1}{
\begin{tabular}{lllllll}
\hline
  Methods & \vline & Cora & Citeseer & Wine & Cancer & Digits\\
  \hline
%   LogReg & \vline & 60.8 (0.0)&62.2 (0.0)& 92.1 (1.3)&93.3 (0.5)&85.5 (1.5)\\
%   Linear SVM & \vline & 58.9 (0.0)&58.3 (0.0)&93.9 (1.6) &90.6 (4.5)& 87.1 (1.8)\\
   RBF SVM & \vline &59.7 (0.0) &60.2 (0.0)&94.1 (2.9) &91.7 (3.1)&86.9 (3.2)\\
%   RF & \vline & 58.7 (0.4)&60.7 (0.7)& 93.7 (1.6)&92.1 (1.7)& 83.1 (2.6)\\
%   FFNN & \vline & 56.1 (1.6)&56.7 (1.7)&89.7 (1.9) &92.9 (1.2)&36.3 (10.3)\\
%   LP & \vline &37.8 (0.2) &23.2 (6.7)&89.8 (3.7) &76.6 (0.5)& 91.9 (3.1)\\
%   ManiReg & \vline &62.3 (0.9) &67.7 (1.6)& 90.5 (0.1)&81.8 (0.1)&83.9 (0.1)\\
   SemiEmb & \vline &63.1 (0.1) &68.1 (0.1)&91.9 (0.1) &89.7 (0.1)& 90.9 (0.1)\\
  LDS & \vline & 84.1 (0.4)&75.0 (0.4)&  97.3 (0.4) & 94.4 (1.9) & 92.5 (0.7)\\
  \hline
  GCN & \vline & 81.0 (0.2)&70.9 (0.3)&\quad\textrm{---} &\quad\textrm{---}&\quad\textrm{---}\\
  GAT & \vline &82.5 (0.4) &70.9 (0.4)&\quad\textrm{---} &\quad\textrm{---}&\quad\textrm{---}\\
  kNN-GCN & \vline & \quad\textrm{---} &\quad\textrm{---}&95.9 (0.9) &94.7 (1.2) &89.5 (1.3) \\
  LDS* & \vline & 83.9 (0.6)& \textbf{74.8 (0.3)}&96.9 (1.4) &93.4 (2.4)&90.8 (2.5)\\
    \hline
  \alg & \vline & \textbf{84.5 (0.3)}&74.1 (0.2)&\textbf{97.8 (0.6)} & \textbf{95.1 (1.0)}& \textbf{93.1 (0.5)}\\
 \hline
\end{tabular}
}
\vspace{-3mm}
\end{table*}

% Inductive setting

\begin{table}[!htb]
% \vspace{-2mm}
\caption{Test scores ($\pm$ standard deviation) in percentage on classification (accuracy) and regression ($R^2$) datasets in the inductive setting.}
\label{table:inductive_results}
\centering
\begin{tabular}{llll}
\hline
  Methods & \vline & 20News & MRD \\
    \hline
  BiLSTM & \vline & 80.0 (0.4) &53.1 (1.4)\\
  kNN-GCN & \vline & 81.3 (0.6)&60.1 (1.5)\\
  \alg & \vline & \textbf{83.6 (0.4)} & \textbf{63.7 (1.8)}\\
 \hline
\end{tabular}
% \vspace{-2mm}
\end{table}

% \subsection{Ablation study}

\begin{table}[!htb]
% \vspace{-2mm}
\caption{Ablation study on various classification datasets.}
\label{table:ablation_results}
\centering
\begin{tabular}{lllllll}
\hline
  Methods & \vline & Cora  & Wine  & 20News\\
  \hline
\alg & \vline & \textbf{84.5 (0.3)}  & \textbf{97.8 (0.6)} & \textbf{83.6 (0.4)}\\
w/o graph reg. & \vline & 84.3 (0.4)&97.3 (0.8) & 83.4 (0.5)\\
w/o IL & \vline &83.5 (0.6) &97.2 (0.8) & 83.0 (0.4)\\
 \hline
\end{tabular}
% \vspace{-4mm}
\end{table}

In this section, we conducted a series of experiments to verify the effectiveness of the proposed model and assess the impact of different model components.  
The implementation of the model will be made publicly available at \url{https://github.com/hugochan/IDGL} soon.
The details on model settings are provided in the appendix.

\subsection{Datasets and Setup}

The benchmarks used in our experiments include 
two network benchmarks, three data point benchmarks and two text benchmarks.
Cora and Citeseer are two commonly used network benchmarks for evaluating graph-based learning algorithms~\citep{sen2008collective}.
The input features are bag of words and the task is node classification. 
In addition to Cora and Citeseer where the graph topology is available, 
we evaluate \alg on three data point benchmarks (i.e., Wine, Breast Cancer (Cancer) and Digits from the UCI machine learning repository~\citep{Dua:2019}).
% For 20News10, we take 10 classes from the 20Newsgroups data and use words (TFIDF) with a frequency of more than 5\% as features.
The task is also node classification.
Finally, to demonstrate the effectiveness of \alg on inductive learning problems, we conduct document classification and regression tasks on the 20Newsgroups data (20News) and the movie review data (MRD)~\citep{pang2004sentimental}, respectively.
In this setting, we regard each document as a graph containing each word as a node.

For Cora and Citeseer, we follow the experimental setup of previous works~\citep{kipf2016semi,velivckovic2017graph,franceschi2019learning}. 
For Wine, Cancer and Digits, we follow the experimental setup of~\citep{franceschi2019learning}.
For 20News, we randomly select 30\% examples from the training data as the development set.
For MRD, we split the data to train/dev/test sets using a 60\%/20\%/20\% split.
The reported results are averaged over 5 runs with different random seeds.
Please refer to
the appendix
for data statistics.

\subsection{Baselines}

Our main baseline in the transductive setting is LDS.
Similar to our work, LDS also jointly learns the graph structure and the parameters of GNNs.
However, LDS is incapable of handling inductive learning problems since it aims at directly optimizing the discrete probability distribution on the edges of the underlying graph, which makes it unable to handle unseen nodes/graphs in the testing phase.
% The experimental results of several semi-supervised (e.g., label propagation (LP)~\citep{zhu2003semi}, manifold regularization (ManiReg)~\citep{belkin2006manifold}, semi-supervised embedding (SemiEmb)~\citep{weston2012deep}) and supervised learning (logistic regression (LogReg), support vector machines (Linear and RBF SVM), random forests (RF), and feed-forward neural networks (FFNN)) baselines are reported in the LDS paper.
The experimental results of several semi-supervised embedding (SemiEmb)~\citep{weston2012deep}) and supervised learning (support vector machines (RBF SVM)) baselines are reported in the LDS paper.
For the sake of completeness, we directly copy their results here.
For ease of comparison, we also copy the reported results of LDS even though we rerun the experiments of LDS using the official code released by the authors.

In addition, for Cora and Citeseer, we include GCN~\citep{kipf2016semi} and GAT~\citep{velivckovic2017graph} as baselines.
In order to evaluate the robustness of \alg to noisy graphs,
we also compare \alg with GCN on graphs with edge deletions or additions.
For data point benchmarks where the graph topology is not available,
we conceive a kNN-GCN baseline where a kNN affinity graph on the data set is first constructed as a preprocessing step before applying a GCN.
For 20News and MRD in the inductive setting, we compare \alg with a BiLSTM~\citep{hochreiter1997long} baseline and kNN-GCN.

\subsection{Results and Analysis}

The results of transductive and inductive experiments are shown in~\cref{table:transductive_results} and~\cref{table:inductive_results}.
First of all, we can see that \alg outperforms all baseline methods in 6 out of 7 benchmarks, which demonstrates the effectiveness of \alg.
% Besides, by comparing the results of GCN, GAT and \alg on Cora and Citeseer, and considering the fact that our method is actually based on GCN, we can conclude that our graph learning method can greatly help the node classification task even when the graph topology is given.
We can clearly see that \alg can greatly help the node classification task even when the graph topology is given.
When the graph topology is not given, 
% we observe that kNN-GCN works well and provides competitive results compared to the supervised baselines that do not leverage graph structures.
% This indicates the benefits of learning and exploiting underlying graph structures.
compared to kNN-GCN, \alg consistently achieves much better results on all datasets, which shows the power of jointly learning graph structures and GNN parameters.
Compared to LDS, \alg achieves better performance in 4 out of 5 benchmarks.
% Unlike LDS which can only handle transductive setting, \alg can easily handle inductive setting without a modification of the algorithm.
% This is because \alg aims at optimizing a metric function instead of the discrete probability distribution on the edges.
The good performance on 20News and MRD verifies the capability  of \alg on inductive learning problems.

We perform an ablation study to assess the impact of different model components.
As shown in~\cref{table:ablation_results}, we can see a significant performance drop consistently on all datasets (e.g., 3.1\% on Citeseer) by turning off the iterative learning component, which demonstrates the effectiveness of the proposed iterative learning framework for the graph learning problem.
We can also see the benefits of jointly training the model with the graph regularization loss.
% For instance, when training the model without the graph regularization loss, the performance on Citeseer drops from 74.1\% to 71.5\%.

\begin{table*}[!htb]
% \vspace{-1mm}
\caption{Mean and standard deviation of training time on various benchmarks (in seconds).}
\label{table:training_time}
\centering
\begin{tabular}{lllllll}
\hline
  Benchmarks & \vline & Cora & Citeseer & Wine& Cancer & Digits\\
  \hline
  GCN & \vline & 3 (1) & 5 (1) & \quad\textrm{---} &\quad\textrm{---} &\quad\textrm{---} \\ 
  GAT & \vline & 26 (5) & 28 (5) & \quad\textrm{---} &\quad\textrm{---} &\quad\textrm{---} \\ 
  LDS & \vline & 390 (82) &585 (181) & 33 (15) &25 (6) & 72 (35)\\ 
  \alg &\vline & 237 (21) & 563 (100) & 20 (7) & 21 (11) & 65 (12)\\ 
  \alg w/o IL &\vline &49 (8) & 61 (15) & 3 (2)  & 3 (1) & 2 (1)\\ 
 \hline
\end{tabular}
% \vspace{-2mm}
\end{table*}

\begin{figure}[!ht]
%   \vspace{-0.1in}
\center
 \includegraphics[keepaspectratio=true,scale=0.12]{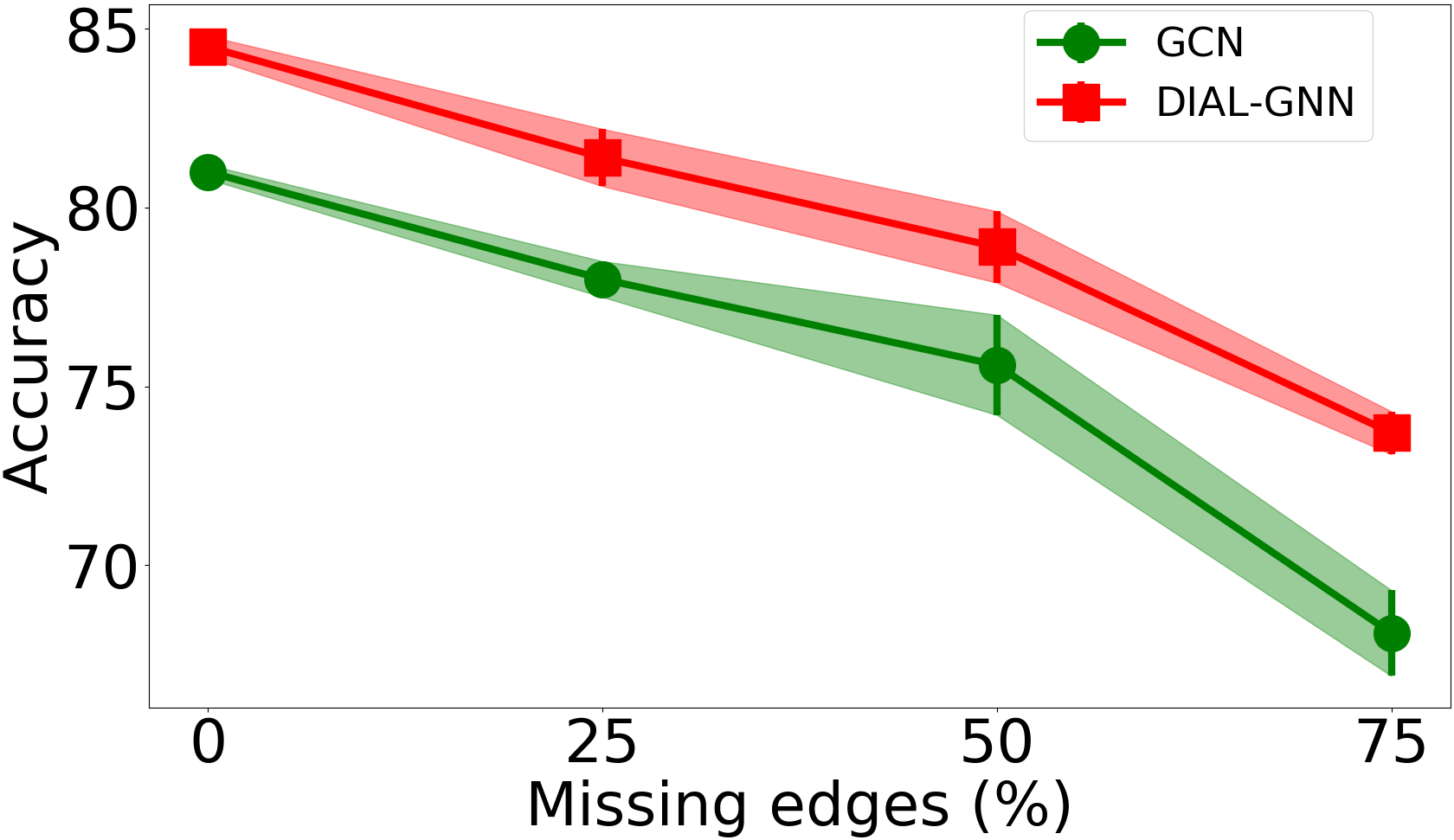}
%   \vspace{-0.1in}
  \caption{Test accuracy ($\pm$ standard deviation) in percentage for the edge deletion scenario on Cora.}
  \label{fig:missing_edges_plot}
%   \vspace{-0.2in}
\end{figure}

\begin{figure}[!ht]
%   \vspace{-0.2in}
\center
\includegraphics[keepaspectratio=true,scale=0.12]{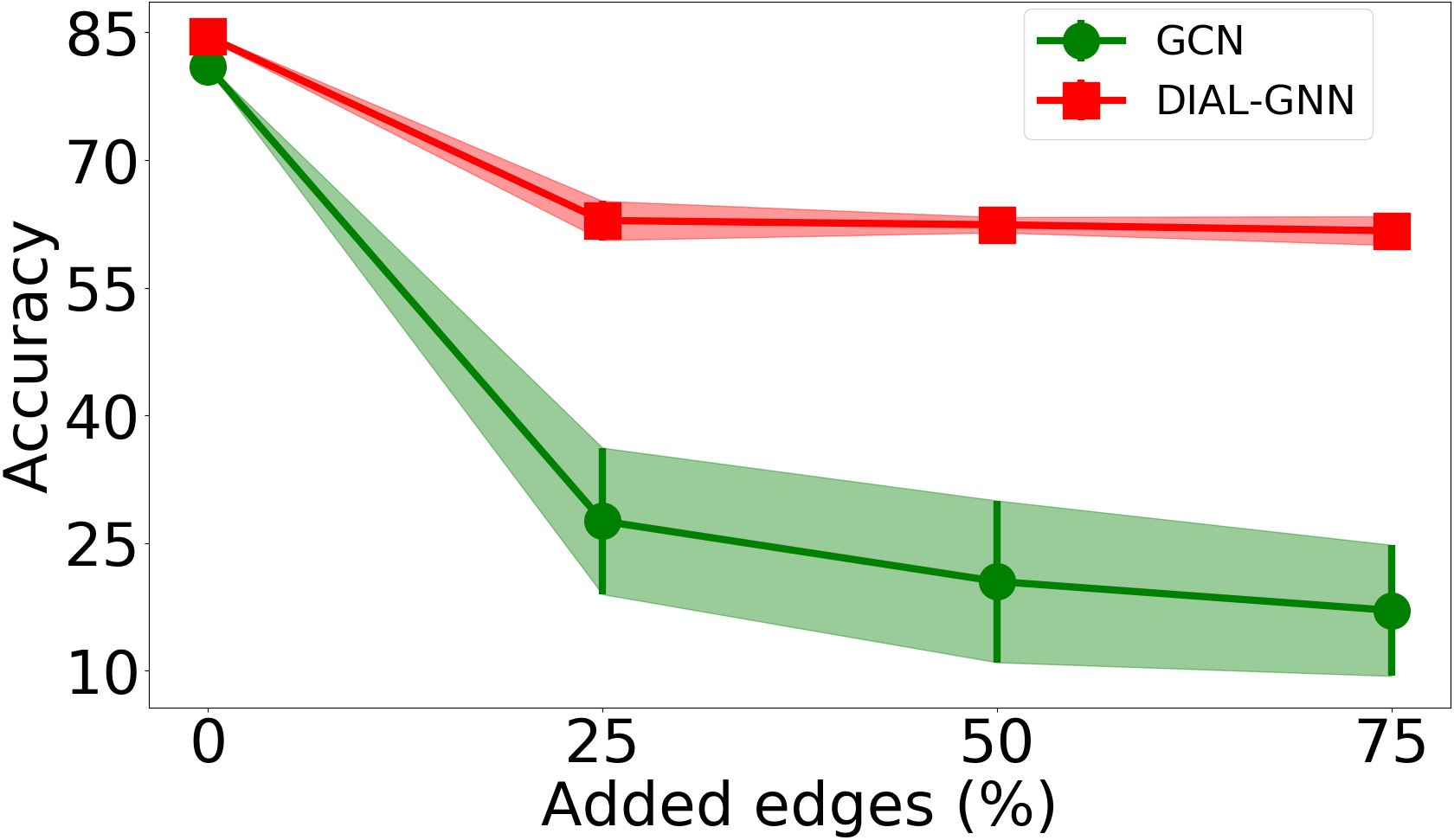}
%   \vspace{-0.1in}
  \caption{Test accuracy ($\pm$ standard deviation) in percentage for the edge addition scenario on Cora.}
  \label{fig:added_edges_plot}
%   \vspace{-0.2in}
\end{figure}

To evaluate the robustness of \alg on noisy graphs, we construct graphs with random edge deletions or additions.
Specifically, we randomly remove or add 25\%, 50\% and 75\% of the edges in the original graphs.
The results on the edge deletion graphs and edge addition graphs are shown in~\cref{fig:missing_edges_plot} and~\cref{fig:added_edges_plot}, respectively.
As we can clearly see, compared to GCN, \alg achieves better results in all scenarios and is much more robust to noisy graphs. 
While GCN completely fails in the edge addition scenario, \alg is still able to perform reasonably well. 
We conjecture this is because~\cref{eq:adj_norm} is formulated in a form of skip-connection, by lowering the value of $\lambda$, we enforce the model to rely less on the initial noisy graph that contains too much additive random noise.

\begin{figure}[!ht]
%   \vspace{-0.2in}
\center
\includegraphics[keepaspectratio=true,scale=0.13]{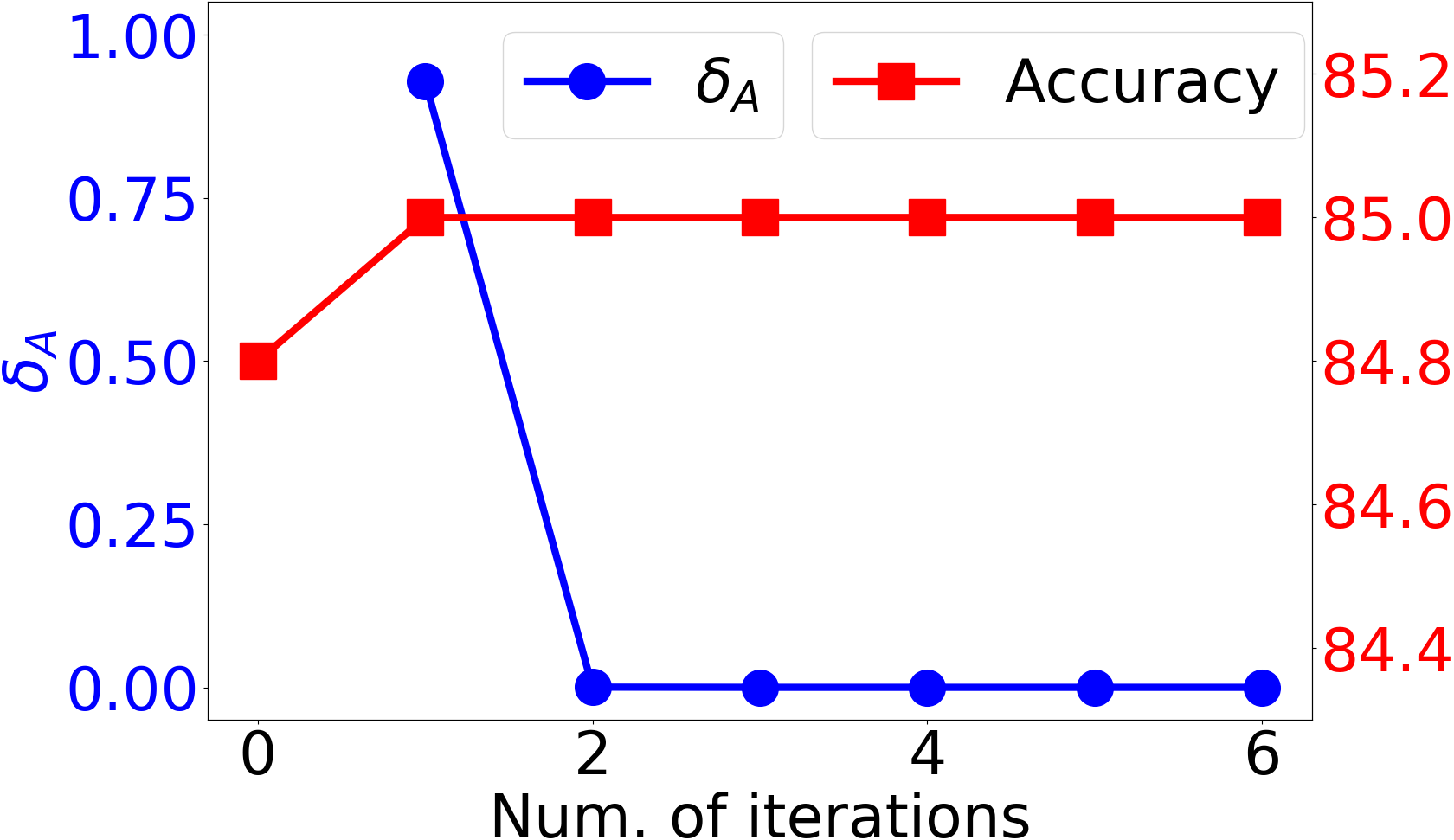}
  \caption{Evolution of the learned adjacency matrix and test accuracy (in \%) through iterations in the iterative learning procedure.}
  \label{fig:convergence}
%   \vspace{-0.2in}
\end{figure}

% \subsection{On the Convergence Property}
In~\cref{fig:convergence}, we show the evolution of the learned adjacency matrix and accuracy through iterations in the iterative learning procedure in the testing phase.
We compute the difference between adjacency matrices at consecutive iterations as $\delta_A^{(t)} = ||\vec{A}^{(t)} - \vec{A}^{(t-1)}||_F^2 / ||\vec{A}^{(t)}||_F^2$ which typically ranges from 0 to 1.
As we can see, both the adjacency matrix and accuracy converge quickly through iterations.
This empirically verifies the analysis we made on the convergence property of the iterative learning procedure.

\begin{figure}[!ht]
%   \vspace{-0.1in}
\center
    \includegraphics[keepaspectratio=true,scale=0.13]{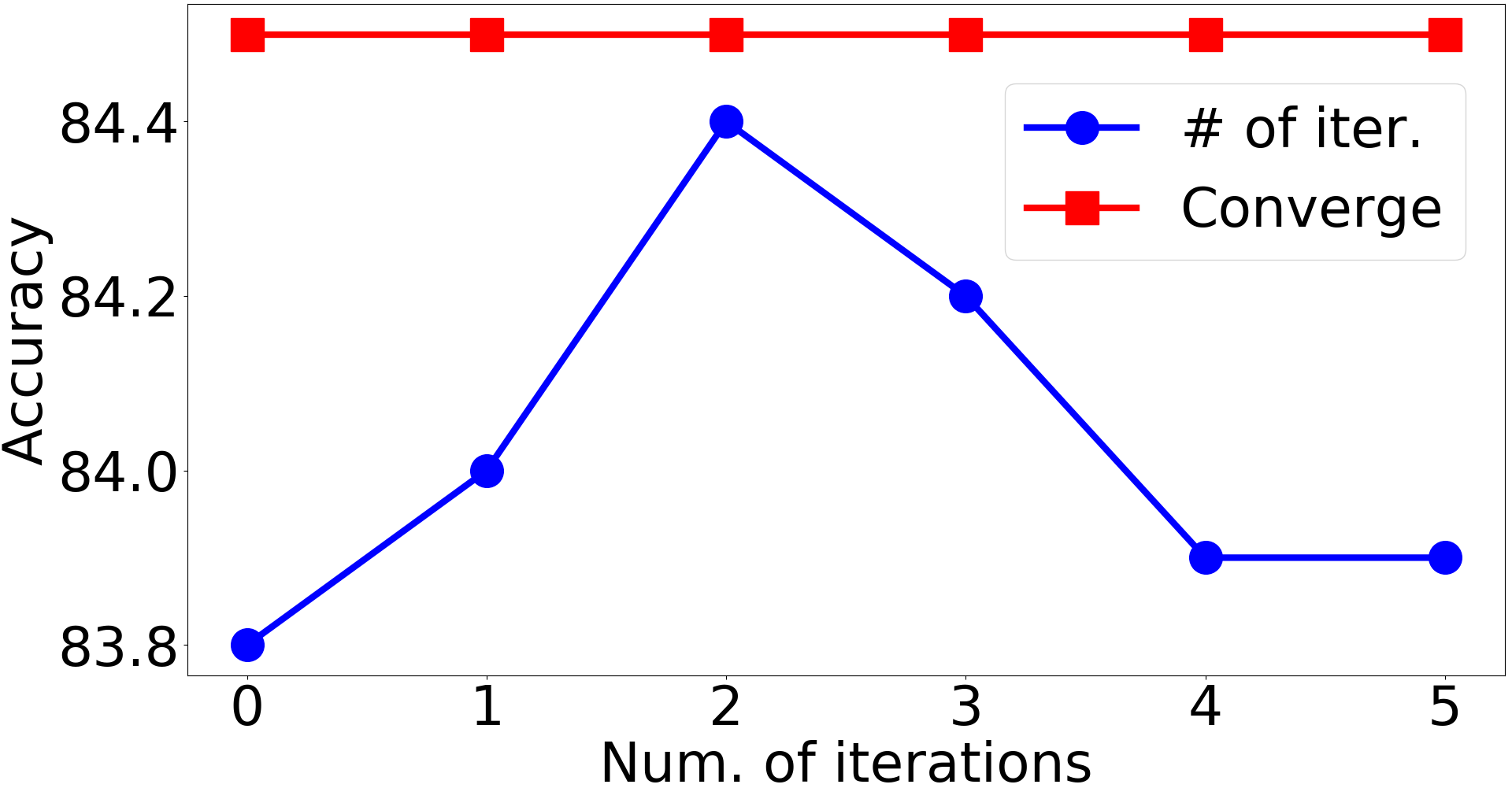}
    \caption{Performance comparison (i.e., test accuracy in \%) of two different stopping strategies: i) using a fixed number of iterations (blue line), and ii) using a stopping criterion to dynamically determine the convergence (red line).}
  \label{fig:effect_stopping}
%   \vspace{-0.2in}
\end{figure}

There are two natural ways of designing the stopping strategy for iterative learning methods.
We can either use a fixed number of iterations, or dynamically determine if the learning procedure already converges or not based on some stopping criterion.
In~\cref{fig:effect_stopping}, we empirically compare the effectiveness of the above two strategies.
We run \alg on Cora (left) and Citeseer (right) using different stopping strategies with 5 runs, and report the average accuracy.
% Note that in all experiments, we keep the stopping strategy as the same in both training and testing phase.
As we can see, dynamically adjusting the number of iterations using the stopping criterion works better in practice.

\subsection{Timing}

Finally, we compare the training efficiency of \alg, LDS and other classic GNNs (e.g., GCN and GAT) on various benchmarks.
All experiments are conducted on the same machine which has an Intel i7-2700K CPU, an Nvidia Titan Xp GPU and 16GB RAM, and are repeated 5 times with different random seeds.
Results are shown in Table~\ref{table:training_time}.
As we can see, both \alg and LDS are slower than GCN and GAT, which is as expected since GCN and GAT do not need to learn graph structures simultaneously.
\alg is consistently faster than LDS, but in general, they are comparable.
We also find that the iterative learning part is the most time consuming in \alg.

\section{Conclusion}

In this paper, we proposed a Deep Iterative and Adaptive Learning framework for Graph Neural Networks (\alg) for 
jointly learning the graph structure and graph embeddings
by optimizing a joint loss combining both task prediction loss and graph regularization loss.
The proposed method is able to iteratively search for hidden graph structures that better help the downstream prediction task.
Our extensive experiments demonstrate that the proposed \alg model can consistently outperform or match state-of-the-art baselines on various datasets.
We leave how to design more effective and scalable metric functions as future work.

\subsubsection*{Acknowledgments}

This work is supported by IBM Research AI through the IBM AI Horizons Network. 
We thank the anonymous reviewers for their helpful feedback.

% \nocite{*}

\bibliography{dgl_aaai20}
\bibliographystyle{aaai}

\newpage
\appendix

\section{Data Statistics}
\label{sec:data_statistics}
The benchmarks used in our experiments include 
two network benchmarks, three data point benchmarks and two text benchmarks.
Below we show the brief data statistics.

\begin{table}[!h]
% \vspace{-4mm}
\caption{Data statistics.}
\label{table:data_statistics}
\centering
\scalebox{0.86}{
\begin{tabular}{lllll}
\hline
  Benchmarks & \vline & Train/Dev/Test&Task&Setting\\
  \hline
  Cora & \vline &140/500/1,000
&node clf & transductive\\ 
Citeseer & \vline &120/500/1,000
&node clf & transductive\\ 
Wine & \vline &10/20/158
&node clf & transductive\\ 
Cancer & \vline &10/20/539
&node clf & transductive\\ 
Digits & \vline &50/100/1,647
&node clf & transductive\\ 
20News & \vline &7,919/3,395/7,532
&graph clf & inductive\\ 
MRD & \vline &3,003/1,001/1,002
&graph reg & inductive\\ 
 \hline
\end{tabular}
}
% \vspace{-4mm}
\end{table}

\section{Model Settings}\label{sec:model_settings}

In all our experiments, we apply a dropout ratio of 0.5 after GCN layers except for the output GCN layer.
During the iterative learning procedure, we also apply a dropout ratio of 0.5 after the intermediate GCN layer, except for Citeseer (no dropout) and Digits (0.3 dropout).
For experiments on text benchmarks,
we keep and fix the 300-dim GloVe vectors for words that appear more than 10 times in the dataset.
For long documents, for the sake of efficiency, we cut the text length to maximal 1,000 words.
We apply a dropout ratio of 0.5 after word embedding layers and BiLSTM layers.
The batch size is set to 16.
And the hidden size is set to 128 and 64 for 20News and MRD, respectively.
For all other benchmarks, the hidden size is set to 16 to follow the original GCN paper.
We use Adam \citep{kingma2014adam} as the optimizer.
For the text benchmarks, we set the learning rate to 1e-3. 
For all other benchmarks, we set the learning rate to 0.01 and apply L2 norm regularization with weight decay set to 5e-4.
\cref{table:hyperparam} shows the hyperparameters associated to \alg for all benchmarks.
All hyperparameters are tuned on the development set.

\begin{table}[!htb]
% \vspace{-4mm}
\caption{Hyperparameter associated to \alg on all benchmarks.}
\label{table:hyperparam}
\centering
\addtolength{\tabcolsep}{-1.8pt}
\scalebox{0.8}{
\begin{tabular}{lllllllllllll}
\hline
  Benchmarks & \vline &$ \lambda $& $\eta$ & $\alpha$& $\beta$ & $\gamma$ & $k$ &$ \epsilon$ &$ m$ & $\delta$ & $ T$\\
  \hline
  Cora & \vline &0.8& 0.1& 0.2 &0.0&0.0 &\textrm{--}&0.0&4&4e-5 & 10\\ 
  Citeseer & \vline &0.6& 0.5& 0.4 &0.0&0.2 &\textrm{--}&0.3&1&1e-3 & 10\\
 Wine & \vline &0.8& 0.7& 0.1 &0.1&0.3 &20&0.75&1&1e-3 & 10\\
 Cancer & \vline &0.25& 0.1& 0.4 &0.2&0.1 &40&0.9&1&1e-3 & 10\\
 Digits & \vline &0.4& 0.1& 0.4 &0.1&0.0 &24&0.65&8&1e-4 & 10\\
 20News & \vline &0.1& 0.4& 0.5 &0.01&0.3 &950&0.3&12&8e-3 & 10\\
 MRD & \vline &0.5& 0.9& 0.2 &0.0&0.1 &350&0.4&5&4e-2 & 10\\
 \hline
\end{tabular}
}
% \vspace{-4mm}
\end{table}

\end{document}